\documentclass[lettersize,journal]{IEEEtran}
\usepackage{amsmath,amsfonts,amssymb}
\usepackage{algorithmic}
\usepackage{algorithm}
\usepackage{array}
\usepackage[caption=false,font=normalsize,labelfont=sf,textfont=sf]{subfig}
\usepackage{textcomp}
\usepackage{stfloats}
\usepackage{url}
\usepackage{verbatim}
\usepackage{graphicx}
\usepackage{cite}
\usepackage{multirow}
\usepackage{booktabs}
\usepackage{pifont}
\usepackage[final]{pdfpages}
\usepackage{xcolor}
\usepackage{soul}

\begin{document}

\title{Context-Aware Timewise VAEs for Real-Time Vehicle Trajectory Prediction}

\author{Pei Xu, Jean-Bernard Hayet, and Ioannis Karamouzas
\thanks{* This work was supported by the National Science Foundation under Grant No. IIS-2047632 and by an unrestricted gift from Roblox. }
\thanks{Pei Xu and Ioannis Karamouzas are with Clemson University, USA. {\tt\small \{peix, ioannis\}@clemson.edu}. Jean-Bernard Hayet is with CIMAT, A.C., Mexico. {\tt\small jbhayet@cimat.mx}. Ioannis Karamouzas is also with University of California, Riverside, USA.}
}



\maketitle

\begin{abstract}
Real-time, accurate prediction of human steering behaviors has wide applications, from developing intelligent traffic systems to deploying autonomous driving systems in both real and simulated worlds. In this paper, we present ContextVAE, a context-aware approach for multi-modal vehicle trajectory prediction. Built upon the backbone architecture of a timewise variational autoencoder, ContextVAE observation encoding employs a dual attention mechanism that accounts for the environmental context and the dynamic agents' states, in a unified way. By utilizing features extracted from semantic maps during agent state encoding, our approach takes into account both the social features exhibited by agents on the scene and the physical environment constraints to generate map-compliant and socially-aware trajectories. We perform extensive testing on the nuScenes prediction challenge~\cite{nuscenes}, Lyft Level 5 dataset~\cite{lyft} and Waymo Open Motion Dataset~\cite{waymo} to show the effectiveness of our approach and its state-of-the-art performance. In all tested datasets, ContextVAE models are fast to train and provide high-quality multi-modal predictions in real-time. Our code is available at: \textnormal{\texttt{https://github.com/xupei0610/ContextVAE}}.
\end{abstract}


\section{Introduction}
Real-time trajectory prediction for traffic agents is fundamental for the development of self-driving systems and intelligent traffic control solutions~\cite{kolekar2021}.
In this paper, we particularly focus on the problem of predicting the multi-modal behavior of vehicles 
in challenging scenes populated with heterogeneous neighbors,
including pedestrians, cyclists and other vehicles. 
Accurate prediction for traffic agents requires the predictive model to fully take into account the external (observable) and internal (intentions) characteristics of the agents, in addition to the contextual influence from the agent's neighbors and the scene environment~\cite{konev2021motioncnn}. 
While numerous approaches have been proposed over the past few years for context-aware vehicle trajectory prediction~\cite{survey},
most existing solutions focus on the model's prediction accuracy
ignoring the requirements for real-time inference performance~\cite{kamenev2022predictionnet}. 
To address this issue, we introduce ContextVAE as a real-time approach for high-fidelity vehicle trajectory prediction.

\begin{figure}[!t]
        \setlength{\fboxsep}{0pt}
        \hfill
    \includegraphics[width=.8\linewidth]{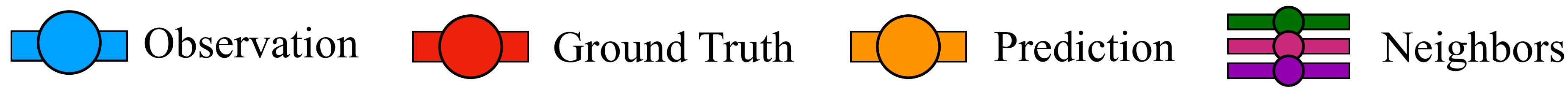}
    
    \centering
    
    \fcolorbox{gray}{white}{\includegraphics[width=.32\linewidth]{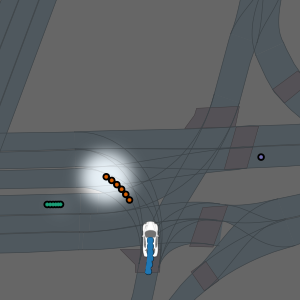}}
    \fcolorbox{gray}{white}{\includegraphics[width=.32\linewidth]{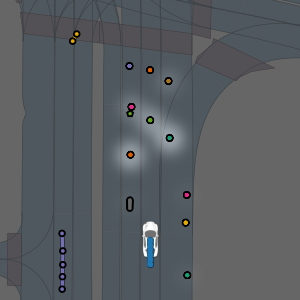}}
    \fcolorbox{gray}{white}{\includegraphics[width=.32\linewidth]{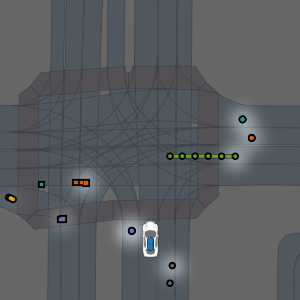}}\\\vspace{0.1cm}

    \fcolorbox{gray}{white}{\includegraphics[width=.32\linewidth]{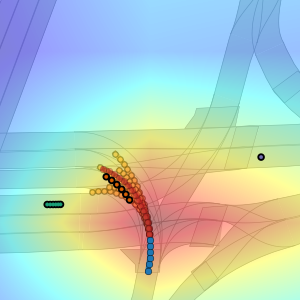}}
    \fcolorbox{gray}{white}{\includegraphics[width=.32\linewidth]{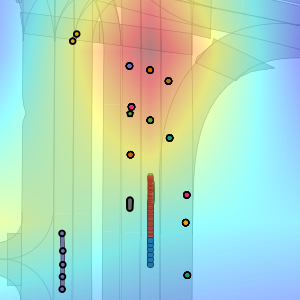}}
    \fcolorbox{gray}{white}{\includegraphics[width=.32\linewidth]{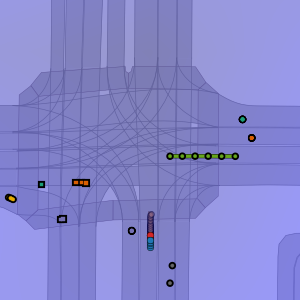}}
    \caption{
    Trajectory predictions on the Lyft dataset using our proposed  approach.  ContextVAE employs a timewise VAE architecture and a context-aware observation  encoding scheme that accounts for  environmental (map) and social (neighbor) features %
    in a unified manner.
    Top: Corresponding attention that the target vehicle is paying to its neighbors within a radius of $30m$.
    Bottom: Predicted trajectories and map regions with high activation in the attention model.
    Red regions are considered more visually salient than blue regions. 
    The white car identifies the vehicle agent under prediction.
   }
    \label{fig:teaser}
\end{figure}

Our work follows the line of recent variational autoencoder (VAE) methods %
that have been shown to achieve state-of-the-art performance in agent trajectory prediction tasks involving homogeneous, human-human interaction settings~\cite{Lee2017,salzmann2020trajectron++,yao2021bitrap}. %
Despite their success, VAE-based methods tend to underperform when tasked to predict vehicle trajectories in complex, heterogeneous traffic scenes~\cite{nuscenes}.
To that end, we explore recent trends in \emph{timewise} VAE architectures~\cite{socialvae2022}
to better capture the highly dynamic and multi-modal nature of agent interactions.  
In addition, we propose a novel, \emph{dual-attention} mechanism  
for observation encoding  
that simultaneously accounts for 
the environmental context (roads, lanes, etc.)  
extracted from the %
local map features 
and the social context 
extracted from the perceived neighbors' states. 
To enable real-time inference, our approach uses an RNN-based encoding of the perceived neighbors~\cite{salzmann2020trajectron++} %
instead of plotting their trajectories on rasterized maps~\cite{kamenev2022predictionnet}. %
Meanwhile, the extracted environmental features directly participate in the agent's state encoding process, leading to a fully, context-aware scheme for observation encoding 
and fast and accurate %
inference of the agent's navigation strategy.

This paper proposes ContextVAE, a real-time approach for 
vehicle trajectory prediction conditioned on short-term observations.
Overall, we make the following contributions: 
\begin{itemize}
\item We introduce a novel, dual-attention architecture for RNN-based observation encoding that accounts for environmental (map) and social (neighbors) features in a \emph{unified} manner.  
We show that such a unified scheme is paramount to achieve high-quality predictions  
compared to the decoupled schemes that current VAE-based solutions employ, where environmental and social features are encoded separately
~\cite{espinoza2022deep,salzmann2020trajectron++,yao2021bitrap,sadeghian2019sophie}.

\item We combine our observation encoding scheme with a VAE architecture where latent variables are sampled timewisely to better model the uncertainty in agent decision making~\cite{socialvae2022}. The resulting ContextVAE approach is generic as it does use any post-processing techniques or prior assumptions about the agents' motion, leading to state-of-the-art performance compared to existing solutions with similar attributes. 
In addition, ContextVAE models are orders of magnitude faster to train due to their low complexity and small memory footprint, and exhibit real-time prediction performance ($<30$ms).

\item We demonstrate the effectiveness and computational efficiency of ContextVAE on three heterogeneous vehicle datasets: the nuScenes prediction challenge~\cite{nuscenes}, Lyft Level 5 dataset~\cite{lyft} and Waymo Open Motion Dataset~\cite{waymo}. 
ContextVAE provides high-fidelity predictions while allowing for real-time inference
and
can serve as a strong baseline for future work
of vehicle trajectory prediction.
\end{itemize}

\section{Related Work}
There has been a lot of prior work on agent trajectory prediction focusing on human-human interactions~\cite{alahi2016social,gupta2018social,Lee2017,sgan2018,sways2019} and autonomous driving applications~\cite{deo2018convolutional,rhinehart2019precog}.
We refer to recent excellent surveys for an overview~\cite{survey,rudenkosurvey}.
Below, we briefly focus on methods exploiting both environmental and social contexts for \emph{vehicle} trajectory prediction.

\emph{Leveraging environmental context: }
To produce map-compliant predictions, early works directly plot the observed trajectories (agent of interest and neighbors) on the rasterized scene maps, performing prediction mainly by using convolutional neural networks (CNNs) to extract map features without fully exploiting the agents' states
~\cite{lyft,kamenev2022predictionnet,phan2020covernet,konev2021motioncnn,chai2019multipath,cui2019mtp}.
More recent works exploit both the environmental context and vectorized neighbors' states as inputs to the prediction model to better 
synthesize 
the agent's motion states and context~\cite{deo2020p2t,autobots2021,zhang2021goalnet,huang2020diversitygan,salzmann2020trajectron++,wang2022stepwise,yao2021bitrap}.
Beyond rasterized maps, some approaches leverage the graph representation of high-definition maps to avoid information loss during rasterization~\cite{gao2020vectornet,liang2020lanegcn,shi2022mtr,gilles2022gohome}.
Systems such as VectorNet~\cite{gao2020vectornet} and AutoBots~\cite{autobots2021} have shown that
CNNs applied on rasterized maps can be replaced by graph neural networks (GNNs) using graph-represented maps.
Instead of modeling the overall environment, QCNet~\cite{qcnet} employs a polygon-based representation ofthe environment and denotes the scene via embedding each agent-polygon pair.

\emph{Generating sequential predictions: } 
CoverNet~\cite{phan2020covernet} and MultiPath~\cite{chai2019multipath} generate trajectories by picking them from a predefined set and cast the generative task as a classification problem.
WIMP~\cite{khandelwal2020wimp}, Trajectron++\cite{salzmann2020trajectron++} and DiversityGAN~\cite{huang2020diversitygan} use recurrent neural networks (RNNs) to perform prediction sequentially.
MTP~\cite{cui2019mtp}, AutoBots~\cite{autobots2021} and LaPred~\cite{kim2021lapred} output fixed-length predicted trajectories at once.
WIMP~\cite{khandelwal2020wimp}, CXX~\cite{luo2020cxx}, TNT~\cite{zhao2021tnt}, GoalNet~\cite{zhang2021goalnet} first predict goal positions and then generate trajectories between the last observed position and the predicted goals. 
IMAP~\cite{espinoza2022deep} and M2I~\cite{sun2022m2i} do not predict trajectories only for target agents but perform prediction sequentially for all the agents.
P2T~\cite{deo2020p2t} and PGP~\cite{deo2021pgp} use reinforcement learning,
with a reward-based model upon which a policy is trained to perform prediction. 

\emph{Handling multi-modality: }
To model the uncertainty during human driving and account for multi-modal decision making, 
Trajectron++\cite{salzmann2020trajectron++} and BiTrap~\cite{yao2021bitrap} follow DESIRE~\cite{Lee2017} and exploit a conditional variational autoencoder (CVAE) architecture where the prior on the latent variable is modeled as a mixture of Gaussians.
MTP~\cite{cui2019mtp}, AutoBots~\cite{autobots2021},
and MultiPath++~\cite{varadarajan2022multipath++} use Gaussian mixtures as well but with an encoder-decoder architecture.
DiversityGAN~\cite{huang2020diversitygan} takes the architecture of generative adversarial networks to produce stochastic results.
LaPred~\cite{kim2021lapred} and LanceGCN~\cite{liang2020lanegcn}, employs $K$ predictor networks with random initialization to produce $K$ predictions parallelly.
Our approach uses a CVAE~\cite{socialvae2022} as its backbone architecture but the latent variables are introduced timewisely~\cite{socialvae2022} to better capture the uncertainty in agent decision making.  
The final distribution is modeled as a sequentially conditioned Gaussian distribution instead of a mixture model.

\emph{Introducing prior knowledge: }
Recent works introduce prior knowledge~\cite{casas2020importance} into the process of vehicle trajectory prediction by assuming that vehicles' possible movement patterns. For example, CXX~\cite{luo2020cxx}, WIMP~\cite{khandelwal2020wimp}, PGP~\cite{deo2021pgp}, LaPred~\cite{kim2021lapred}, LaneRCNN~\cite{zeng2021lanercnn}, HOME~\cite{gilles2021home}
and GOHOME~\cite{gilles2022gohome} select interesting paths traversing the road graph or specific lanes in the graph as the basis to generate numeric coordinates. 
TNT~\cite{zhao2021tnt} and GoalNet~\cite{zhang2021goalnet} select goal positions based on the road graph and then perform goal-directed prediction.
Though achieving impressive results,
such approaches heavily rely on the assumption
that vehicles will move along the road graph,
which is not always true in real-life scenarios. 
Thus, we focus on comparisons with works that do not rely on this assumption.

\section{Approach}\label{sec:approach}
\begin{figure*}[!t]
    \centering
    \includegraphics[width=.775\linewidth]{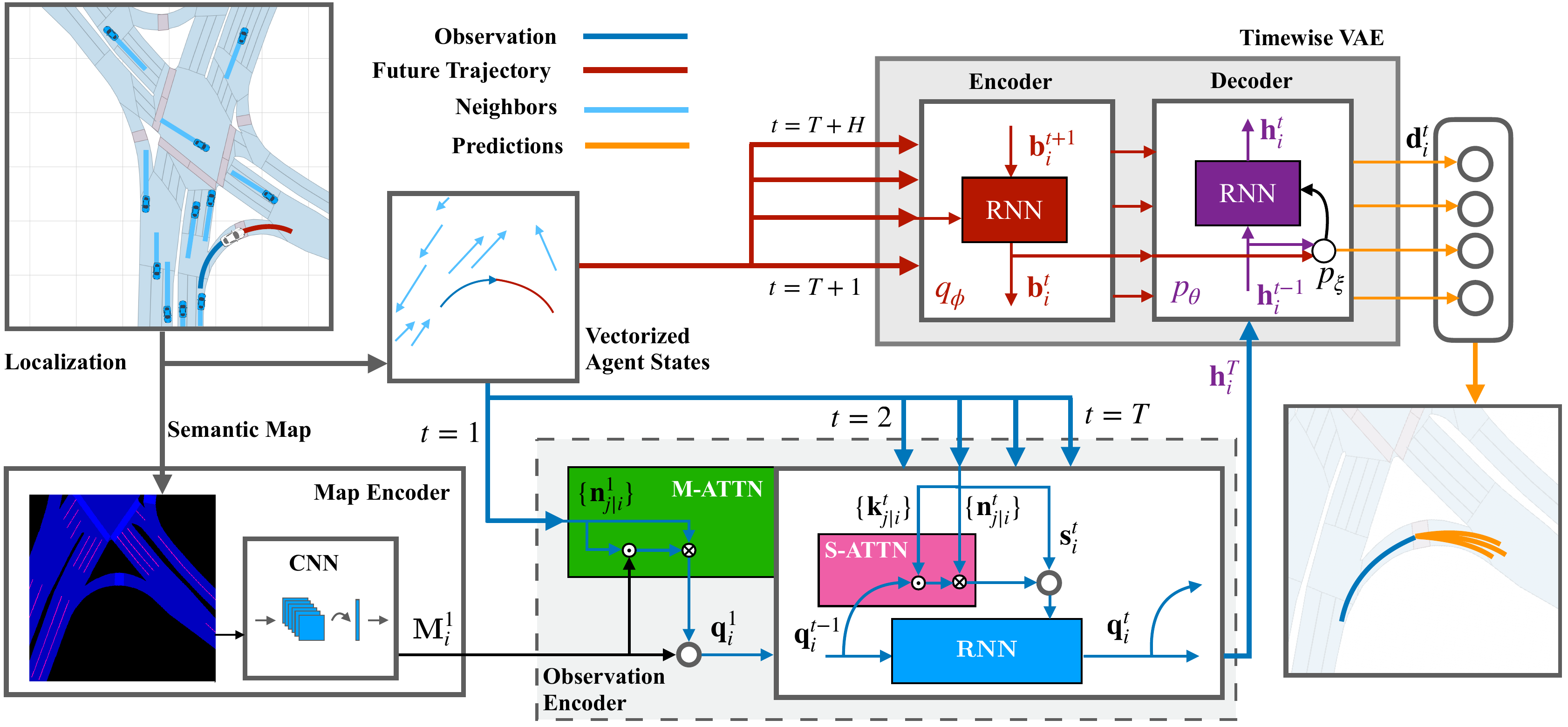}
    \caption{%
    ContextVAE backbone architecture is a timewise VAE shown in the gray block at the upper right.
    The proposed map encoding module is linked to the backbone timewise VAE architecture through the context-aware observation encoder~(dashed box), which processes the vectorized agent states in a unified way with the map features. 
    M-ATTN and S-ATTN represent the map and social attention respectively, where
    $\odot$ is the dot product operator,
    $\otimes$ is the element-wise multiplication operator,
    and $\bigcirc$ is the concatenation operator. Red parts for posterior estimation are used only during training.}
    \label{fig:overview}
\end{figure*}

\subsection{Problem Formulation}
Given a local, $T$-frame observation $\mathcal{O}_i^{1:T}$ gathered from a target agent $i$,
we seek  to estimate the agent's trajectory predictive distribution over the future $H$ frames, i.e. $p(\mathbf{x}_i^{T+1:T+H} \vert \mathcal{O}_i^{1:T})$, where $\mathbf{x}_i^{T+1:T+H}$ are the agent's future coordinates. 
To perform context-aware prediction, we use observations that include both the $N_i$ observed neighboring agent positions and the local environmental context information $\mathbf{M}_i^t$, i.e.
\mbox{$\mathcal{O}_i^t := \{\mathbf M_i^t, \{\mathbf{x}_j^t\}_j\}$}, 
where $j = 1,\cdots,N_i$ 
and $\vert\vert \mathbf{x}_j^t - \mathbf{x}_i^t \vert\vert \leq r_i$ given $r_i$ as the observation radius of agent $i$.
We model $\mathbf{M}_i^t$
as a vector of features extracted from rasterized semantic maps via a CNN module, as shown in Fig.~\ref{fig:overview}.
To train our models in a scene-invariant way, 
we do not predict directly the global coordinates
$\mathbf{x}_i^{T+1:T+H}$ but 
rather the \emph{displacement} distribution $p(\mathbf{d}_i^{T+1:T+H} \vert \mathcal{O}_i^{1:T})$,
where $\mathbf{d}_i^t := \mathbf{x}_i^t-\mathbf{x}_i^{t-1}$ is the displacement between two frames.
Future trajectories are reconstructed as $\mathbf{x}_i^{T+\tau} = \mathbf{x}_i^T + \sum_{t=T+1}^{T+\tau} \mathbf{d}_i^\tau$ by sampling $\mathbf{d}_i^\tau$ from the predicted distributions.%

\subsection{Model Architecture and Training Objective}
\label{sec:backbone}
The backbone architecture of ContextVAE is a timewise VAE with 
a conditional prior $p_\theta$
and a posterior $q_\phi$
estimated bidirectionally (see Fig.~\ref{fig:overview}). 
The training objective is to maximize a timewise evidence lower bound: 
\begin{equation}\label{eq:elbo_v}\begin{split}
    \mathbb{E}_i \Biggl[ \frac{1}{H}
    \sum_{t=T+1}^{T+H}&\mathbb{E}_{\mathbf{z}_{i}^t \sim q_\phi(\cdot \vert \mathbf{b}_{i}^t, \mathbf{h}_{i}^{t-1})}\biggl[ \log p_\xi(\mathbf{d}_i^t \vert \mathbf{z}_i^t,\mathbf{h}_i^{t-1}) \\
    & - D_{KL}\left[q_\phi(\mathbf{z}_i^t \vert \mathbf{b}_i^t, \mathbf{h}_i^{t-1}) \vert\vert p_\theta(\mathbf{z}_i^t \vert \mathbf{h}_i^{t-1})\right]\biggr]
    \Biggr],
\end{split}\end{equation}
where $\mathbf{z}_i^t$ is the latent variable sampled timewisely, and
$\mathbf{b}_i^t$ and $\mathbf{h}_i^{t-1}$ are the RNN hidden states, which are updated backward in the encoder and forward in the decoder, respectively.
We parameterize the output distribution $p_\xi$, the posterior $q_\phi$ and the prior $p_\theta$ as Gaussians, using neural networks. 

The timewise VAE decoder hidden state $\mathbf{h}_i^{t}$ for $t>T$ is updated recurrently based on the sampled $\mathbf{z}_i^t$ and the displacement $\mathbf{d}_i^t$, i.e. 
\mbox{$\mathbf{h}_i^t = \overrightarrow{g}(\psi_\mathbf{zd}(\mathbf{z}_i^t, \mathbf{d}_i^t), \mathbf{h}_i^{t-1})$}.
While feeding $\mathbf{z}_i^t$ allows to produce the necessary variations to match the predictive distribution, the displacement $\mathbf{d}_i^t$ allows to enforce some continuity in the produced displacements across time.
The initial state $\mathbf{h}_i^T = \psi_\mathbf{h}(\mathcal{O}_i^{1:T})$ encodes the observation,
where $\psi_\mathbf{zd}$ and $\psi_\mathbf{h}$ are embedding networks. 
This leads to a prior conditioned by the context observation $\mathcal{O}_i^{1:T}$ (see~\ref{sec:encoding}).
The timewise VAE encoder hidden state $\mathbf{b}_i^t$ is updated backwards, i.e.
\mbox{$\mathbf{b}_i^t = \overleftarrow{g}(\mathcal{O}_{i}^t,\mathbf{b}_i^{t+1})$},
given the ground-truth observation $\mathcal{O}_{i}^t$ at the future frame %
$t>T$ and a zero initial state $\mathbf{b}_i^{T+H+1}$. %

At training, the posterior $q_\phi$ over the latent variable $\mathbf z_i^t$ at timestep $t$ uses the information collected from the \emph{future} part of the ground-truth trajectory from $T+H$ down to $t$, through the backward encoding $\mathbf b_i^t$. 
This leverages the whole trajectory availability at training time for better extracting the agent's strategy. 
It also uses the forward encoding $\mathbf{h}_i^{t-1}$ evaluated from $1$ to $t-1$, thus considering the \emph{whole} trajectory $\mathcal{O}_i^{1:T+H}$. 
The displacement distribution $p_\xi(\mathbf d)$ relies on this latent variable $\mathbf{z}_t$, drawn from $q_\phi(\cdot \vert \mathbf{b}_i^{t},\mathbf{h}_i^{t-1})$ at training, and on $\mathbf{h}_i^{t-1}$.

At inference, predictions are generated through latents drawn from the prior:
$\mathbf{d}_i^t \sim p_\xi(\cdot\vert\mathbf{z}_i^t,\mathbf{h}_i^{t-1})$
where $\mathbf{z}_i^t \sim p_\theta(\cdot \vert \mathbf{h}_i^{t-1})$,
 relying on the historical observation $\mathcal{O}_i^{1:T}$ only.

\subsection{Context-Aware Observation
Encoding}
\label{sec:encoding}
We introduce a \emph{unified} scheme 
for observation encoding, where the environmental context extracted from the semantic map and the social context extracted from the observed agent states are exploited simultaneously. 
Our scheme leverages a dual attention mechanism 
that combines \emph{map attention} with \emph{social attention} as shown at the bottom of Fig.~\ref{fig:overview}, and is based on the intuition that humans account for environmental features dynamically and instantaneously during driving, rather than encoding these features one after the other.

Following prior literature~\cite{salzmann2020trajectron++,yao2021bitrap,socialvae2022,sways2019}, our encoder employs an RNN structure $g$ (blue box in Fig.~\ref{fig:overview}) that encodes the agent state sequentially. 
The RNN is updated as $\mathbf{q}_i^{t+1} = g(\mathcal{O}_i^t, \mathbf{q}_i^t)$,
where, from the local observation $\mathcal{O}_i^t$, we use the agent's states $\mathbf{s}_i^t$, including its velocity and acceleration, and 
the observed neighbors' state $\{\mathbf{n}_{j \vert i}^t\}$, including
the relative position and velocity of each neighbor $j$ to the agent $i$.
The final $\mathbf{q}_i^T$ is used to initialize the decoder's hidden states in the timewise VAE backbone, %
i.e., 
$\psi_\mathbf{h}(\mathcal{O}_i^{1:T}) \equiv \mathbf{q}_i^T$
(Section~\ref{sec:backbone}). 

To account for how the environmental context influences the agent's steering actions, 
we initialize the hidden state $\mathbf{q}_i^1$ 
using features %
extracted from the local semantic map, rather than encoding the map and the agent’s state 
independently. 
However, instead of directly plugging the map features into the observation encoder,
we introduce a \emph{map attention} (M-ATTN) mechanism and initialize the encoder via
\begin{equation}
\label{eq:q1}
\mathbf{q}_i^1 = \textsc{Concat}\left(\mathbf M_i^1, \text{M-ATTN}(\{\mathbf{n}_{j\vert i}^1\})\right),
\end{equation}
where
\begin{equation}\label{eq:m_attn}\begin{split}
    \text{M-ATTN}(\{\mathbf{n}_{j\vert i}^1\}) &\triangleq \textsc{Attn}(\mathbf{M}_i^1, \{\mathbf{n}_{j\vert i}^1\}, \{\mathbf{n}_{j\vert i}^1\}) \\
    &= \sum\nolimits_j w_{j \vert i} f_{\mathbf{n}_{\text{val}}^1}(\mathbf{n}_{j\vert i}^1).
\end{split}\end{equation}
Here, $\mathbf{M}_i^1$ denotes vectorized features extracted from agent $i$'s local map at the \emph{first} observation frame, and
$f_\text{\Large$\cdot$}$ are embedding networks. %
$\mathbf{M}_i^1$, $\{\mathbf{n}_{j\vert i}^1\}$ and $\{\mathbf{n}_{j\vert i}^1\}$
are the query, key, and value vectors, respectively, of the attention mechanism. 
The attention weights $w_{j \vert i}$ are obtained through dot-product operations between $\mathbf{M}_i^1$ and $\mathbf{n}_{j\vert i}^1$ followed by a softmax operation, i.e. $w_{j \vert i} =\textsc{Softmax}(f_{\mathbf{M}^1}(\mathbf{M}_i^1) \odot f_{\mathbf{n}_{\text{key}}^1}(\mathbf{n}_{j\vert i}^1))$.
This M-ATTN mechanism %
can help identify neighbors close to the target agent but with less influence in the given context, e.g., vehicles spatially close to the %
agent but on a lane that is inaccessible to it. %
We refer to  Section~\ref{sec:case_study} and Fig.~\ref{fig:teaser} for qualitative results and to Section~\ref{sec:sensitivity} for quantitative analysis. 

During agent state encoding, we 
synthesize the social context represented by the neighbors' states $\{\mathbf{n}_{j \vert i}^t\}$ via a \emph{social attention} (S-ATTN) mechanism as: 
\begin{equation}\label{eq:s_attn}\begin{split}
   \textsc{S-ATTN}(\{\mathbf{n}_{j \vert i}^t\}) &\triangleq \textsc{Attn}(\mathbf{q}_i^t, \{\mathbf{k}_{j\vert i}^t\}, \{\mathbf{n}_{j\vert i}^t\}) \\
    &= \sum\nolimits_j \omega_{j \vert i} f_{\mathbf{n}_{\text{val}}}(\mathbf{n}_{j\vert i}^t),
\end{split}\end{equation}
where, 
$\mathbf{q}_i^t$, $\{\mathbf{k}_{j\vert i}^t\}$, and $\{\mathbf{n}_{j\vert i}^t\}$ correspond to the query, key and value vectors of the attention, and each weight $\omega_{j \vert i}$ of the attention graph is computed as the dot product between $\mathbf{q}_i^t$ and $\mathbf{k}_{j\vert i}^t$ after obtaining their fixed-length embeddings.  
Here, $\mathbf{k}_{j\vert i}^t$ denotes the social features exhibited by the neighbor $j$ and observed by the agent $i$. 
As introduced in~\cite{socialvae2022}, we include the distance, bearing angle and minimal predicted distance from agent $i$ to $j$ as the social features $\mathbf{k}_{j\vert i}^t$.

By choosing a proper size of the local map, 
we can have $\mathbf{M}_i^1$ covering the whole trajectory of the agent during the observation and prediction horizon from $t=1$ to $t=T+H$. 
This allows us to sidestep introducing $\mathbf{M}_i^t$ timewisely, and update the hidden state in  the RNN $g$ by representing  $\mathcal{O}_i^t$ as
    $\mathcal{O}_i^t \equiv \textsc{Concat}\left(\mathbf{s}_i^t, \textsc{S-ATTN}(\{\mathbf{n}_{j \vert i}^t\})\right)$.
This way,
we 
perform fast observation encoding directly on the environmental and social contexts 
after one pass of the map encoding module,
and avoid the computationally expensive processing a sequence of rasterized maps.
We refer to Section~\ref{sec:sensitivity} for our model's real-time prediction performance.

As shown in Fig.~\ref{fig:overview}, we exploit rasterized semantic maps to capture contextual information 
and extract the related features $\mathbf M_i^1$ using a CNN module. %
Note that
there is no constraint about the form of $\mathbf M_i^1$:
It can also be represented %
using graphs processed by a GNN~\cite{gao2020vectornet,liang2020lanegcn,zhao2021tnt,deo2021pgp,gilles2022gohome}, 
or using a lane attention block~\cite{luo2020cxx,kim2021lapred}.
Since rasterization is the common way to generate semantic maps for contextual information and since most open-source vehicle datasets %
do not provide an API to directly generate road graphs, in our experiments 
we rely on 
CNNs to extract contextual features from rasterized maps. 

\section{Experiments}
We evaluate our approach on vehicle trajectory prediction tasks using
the nuScenes prediction challenge (\textit{nuScenes})~\cite{nuscenes}, 
Lyft Level 5 Prediction Dataset (\textit{Lyft})~\cite{lyft} and 
Waymo Open Dataset (\textit{Waymo})~\cite{waymo}.
These datasets handle heterogeneous types of agents, including various vehicles, cyclists, and pedestrians. 
We focus our prediction on vehicles' trajectories and all other types of agents are considered as neighbors.

\subsection{Setup}
\label{sec:exp_setup}
\begin{figure}[!t]
    \includegraphics[width=.325\linewidth]{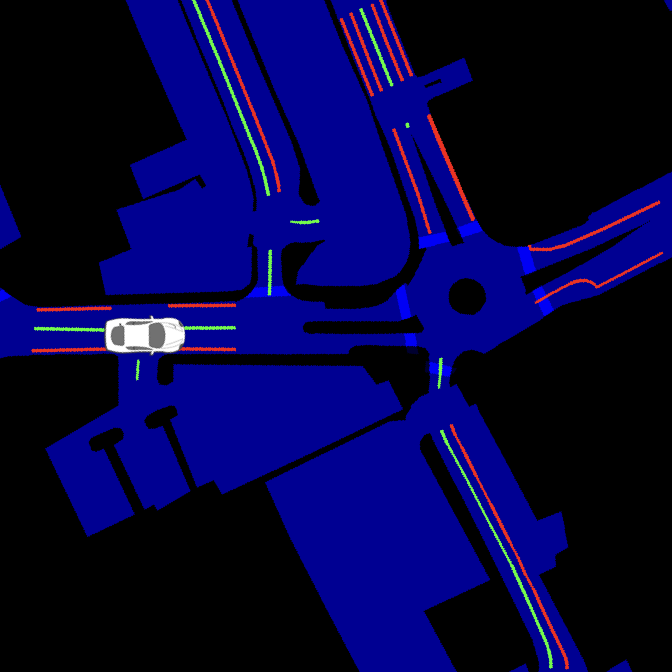}
    \hfill  \includegraphics[width=.325\linewidth]{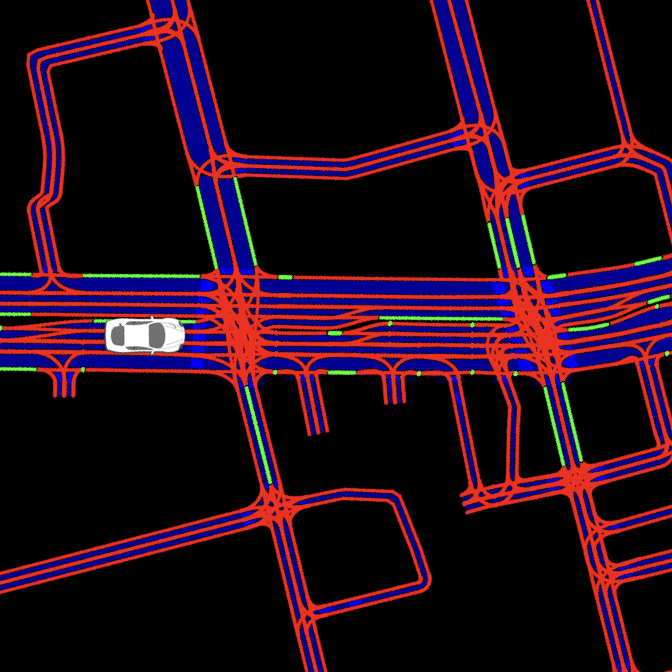}
    \hfill    \includegraphics[width=.325\linewidth]{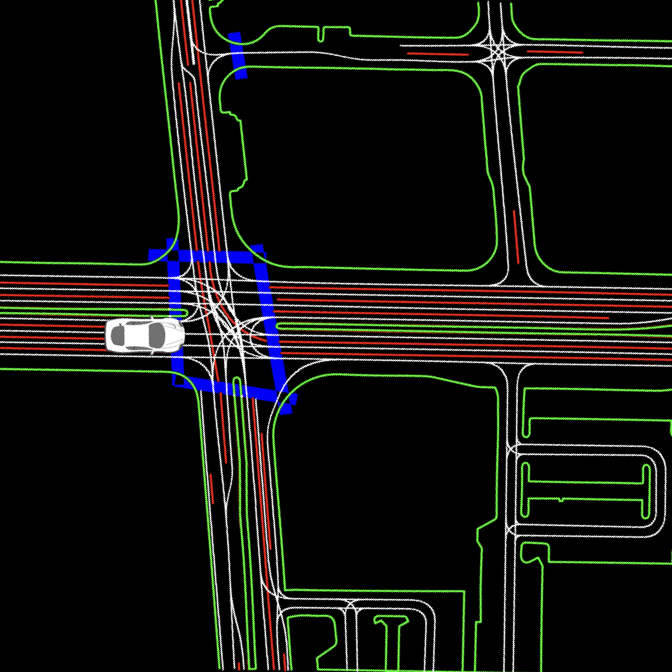}
    
    \caption{Examples of rasterized semantic maps for (from left to right): \textit{nuScenes}, \textit{Lyft} and \textit{Waymo}. Blue identifies drivable areas and crosswalks; green identifies road edges; red identifies lane dividers. 
    The drawn cars identify the first observed position of the agent under prediction and give the map local origin and $x$-axis.
    The cars are not part of the rasterized maps.}
    \label{fig:map_example}
\end{figure}
\paragraph{Implementation details}
We use a rasterized, $224{\times}224$ local semantic map to represent the environmental context.
The map is translated and rotated such that
the vehicle under prediction heads towards the positive $x$-axis and 
stands at the 122nd row and 51st column of the map in the 1st frame of observation.
To be consistent,
the coordinates of all agents are expressed using the same local system.
Figure~\ref{fig:map_example} shows examples of the extracted maps for the three datasets.
Since each dataset has its own semantic map definition, 
the generated rasterized maps are slightly different.
We refer to the supplementary material for details on semantic map rasterization.

\paragraph{Evaluation Metrics and Baselines} 
We use the minimum average displacement error over $k$ predictions (minADE$_k$) and the minimum final displacement error over $k$ predictions (minFDE$_k$) 
to assess a model's performance.
We consider both
the most likely prediction ($k=1$) and the top-5 predictions ($k=5$)   
as key metrics (in meters).  
We provide comparisons to three deterministic baselines: 
\textit{Constant Velocity}%
assumes 
target vehicles %
always move at the velocity defined between the last two observed frames; 
\textit{Constant Lane} assumes target vehicles keep their last observed speed but always along the lane where they were at the last observed frame;
\textit{Kalman Filter} applies an extended Kalman filter while assuming constant longitudinal speed and heading direction.
We also consider state-of-the-art, data-driven baselines for each dataset. 
As ContextVAE does not use prior knowledge to constrain the movement of vehicles,
we limit our comparisons to approaches that do not restrict the accessibility of vehicles via maps or refine the predicted trajectories by post-processing.

\subsection{Quantitative Results}\label{sec:quant_res}
\paragraph{nuScenes} 
We use the official training and validation sets of \textit{nuScenes} prediction challenge to perform model 
training and performance evaluation. 
An observation window of up to $2\,$s is used
and the prediction horizon is $6\,$s. As 
the data was recorded at $2\,$FPS, this leads to a 12-frame prediction and a varying observation 
window between 2 and 5 frames.
Additionally to the heuristic baseline models, %
we introduce three other baselines for comparison. 
The Trajectron++~\cite{salzmann2020trajectron++} model is trained from scratch with two known bugs fixed as indicated in the official repository. 
P2T~\cite{deo2020p2t} is tested using their provided pre-trained model.
Since no pre-trained models are provided, we train the AutoBots-Ego model using the official implementation.  
These three baselines exploit CNN modules for map encoding. 
We report the results in Table~\ref{tab:nuscenes}.
Among the considered baselines, \textit{Constant Velocity}, \textit{Constant Lane} and \textit{Kalman Filter} perform the worst.
While \textit{Constant Lane} ensures that the predicted position is always along a lane, it exhibits worse performance than \textit{Constant Velocity} as it ignores the possibility that the target vehicle would change lanes.
As recently published works, P2T and AutoBots-Ego work significantly better than Trajectron++, with AutoBots-Ego outperforming P2T.  
Our approach not only achieves the best performance on all four metrics, 
but is also very fast to train (see Section~\ref{sec:cnns}).%

\begin{table}[t]
\begin{center}
\caption{Performance on nuScenes Prediction Challenge.}\vspace{-0.2cm}
\label{tab:nuscenes}
\begin{tabular}{l cc cc}
    \toprule
    \multirow{2}{*}{\textit{nuScenes} (Validation Set)} & \multicolumn{2}{c}{minADE$_{k}$ (m)} & \multicolumn{2}{c}{minFDE$_{k}$ (m)} \\
     & $k=1$ & $k=5$ & $k=1$ & $k=5$ \\
    \midrule
    Constant Velocity &  4.61 & - & 11.21 & - \\
    Constant Lane & 5.45 & - & 12.73 & - \\
    Kalman Filter & 4.17 & - & 10.99 & - \\
    Trajectron++ & 4.08 & 2.41 & 9.67 & 5.63  \\
    P2T & 3.82 & 1.86 & 8.95 & 4.08 \\
    AutoBots-Ego & 3.86 & 1.70 & 8.89 & 3.40 \\
    \midrule
    ContextVAE
    & \textbf{3.54} & \textbf{1.59} & \textbf{8.24} & \textbf{3.28} \\
    \bottomrule
\end{tabular}
\end{center}
\end{table}

\begin{table}[t]
\begin{center}
\caption{Performance on Lyft Level 5 Prediction Dataset.}\vspace{-0.2cm}
\label{tab:lyft}
\begin{tabular}{l cc cc}
    \toprule
    \multirow{2}{*}{\textit{Lyft}} & \multicolumn{2}{c}{minADE$_{k}$ (m)} & \multicolumn{2}{c}{minFDE$_{k}$ (m)} \\
     & $k=1$ & $k=5$ & $k=1$ & $k=5$ \\
    \midrule
    Constant Velocity & 0.70 & - & 1.64 & - \\
    {Constant Lane} & {0.78} & - & {2.04} & - \\
    Kalman Filter & 0.60 & - & 1.30 & - \\
    SAMPP~\cite{espinoza2022deep} & 0.39 & - & 0.83 & - \\
    IMAP~\cite{espinoza2022deep} & 0.28 & - & 0.65 & -\\
    Trajectron++ & 0.38 & 0.26 & 0.89 & 0.57 \\
    \midrule
    ContextVAE & \textbf{0.24} & \textbf{0.16} & \textbf{0.54} & \textbf{0.32} \\
    \bottomrule
\end{tabular}
\vspace{-0.7cm}
\end{center}
\end{table}

\begin{table}[t]
\begin{center}
\caption{Performance on Waymo Open Dataset.}\vspace{-0.2cm}
\label{tab:waymo}
\begin{tabular}{l cc cc}
    \toprule
    \multirow{2}{*}{\textit{Waymo} (Validation Set)} & \multicolumn{2}{c}{minADE$_{k}$ (m)} & \multicolumn{2}{c}{minFDE$_{k}$ (m)} \\
     & $k=1$ & $k=5$ & $k=1$ & $k=5$ \\
    \midrule
    Constant Velocity & 2.04 & - & 5.25 & -\\
    {Constant Lane} & {2.54} & - & {5.85} & - \\
    Kalman Filter & 1.99 & - & 4.07 & - \\
    SAMPP~\cite{espinoza2022deep} & 1.26 & - & 2.80 & - \\
    IMAP~\cite{espinoza2022deep} & 0.97 & - & 2.03 & -\\
    Trajectron++ & 0.88 & 0.56 & 2.37 & 1.41 \\
    MotionCNN & 0.83 & 0.40 & 1.99 & 0.81 \\
    M2I & 0.67 & 0.42 & 1.60 & 0.85 \\
    \midrule
    ContextVAE & \textbf{0.59} & \textbf{0.30} & \textbf{1.49} & \textbf{0.68} \\
    \bottomrule
\end{tabular}
\vspace{-0.6cm}
\end{center}
\end{table}

\paragraph{Lyft}
This dataset is much larger than \emph{nuScenes} and has more than 1,000 hours of data with 170,000 scenes. 
We use only the first 16,265 training scenes, each scene being about 24s long. 
For evaluation, we use the full validation dataset (16,220 scenes).
We downsample the data from $10\,$FPS to $5\,$FPS and train models with an observation window of $1\,$s and prediction horizon of $3\,$s.
In Table~\ref{tab:lyft}, we report the performance obtained by our approach and two new baselines: %
SAMPP and IMAP~\cite{espinoza2022deep}.
These are deterministic models that rely on lane graphs rather than rasterized semantic maps.
Trajectron++ is introduced as a multimodal baseline, trained using the same data as ours.
As can be seen, IMAP achieves better deterministic results 
(minADE$_1$/minFDE$_1$) compared to Trajectron++, 
though these results are slightly worse compared to Trajectron++'s multimodal performance.
Our approach outperforms IMAP and Trajectron++ on both the deterministic and multimodal results,
with an improvement around 15\% on minADE$_1$/minFDE$_1$ and around 40\% on minADE$_5$/minFDE$_5$.

\paragraph{Waymo}
This dataset 
has two types of training data formats, consisting of $20\,$s and $9\,$s sequential data, respectively. 
We use the $9\,$s data for training, which contains 487,002 scenes, while the validation set has 44,097 scenes.
In addition to performing downsampling from $10\,$FPS to $5\,$FPS, we apply the data filter from the \textit{Lyft} dataset to filter out invalid data. 
Similar to our \textit{Lyft} implementation 
we use a $1\,$s window for observation and a $3\,$s horizon for prediction.
Table~\ref{tab:waymo} shows the performance of our approach along with two new baselines: MotionCNN~\cite{konev2021motioncnn} and M2I~\cite{sun2022m2i}.
We run the pre-trained models from the two baselines and keep the first 3s prediction for evaluation.
As shown in Table~\ref{tab:waymo}, MotionCNN and M2I outperform Trajectron++ significantly on minADE$_5$/minFDE$_5$.
M2I also brings a large improvement on the most likely prediction. %
Compared with these strong baselines, ContextVAE
achieves more than 15\% improvement overall, and a 29\% and 20\% improvement on minADE$_5$ and minFDE$_5$, respectively.

The default Waymo motion prediction challenge setup uses an 8s prediction horizon with $k=6$.
The performance of our model trained for 8s %
on the \textit{interactive} validation set for vehicle-only target agents is: minADE$_6 = 1.59$ and minFDE$_6 = 3.67$. 
As a comparison, the Waymo LSTM baseline achieves minFDE$_6 = 6.07$ for 8s prediction~\cite{ngiam2021scene}, and M2I~\cite{sun2022m2i} reports minFDE$_6 = 5.49$.

\begin{table*}
\begin{center}
\caption{Prediction errors with Different CNN Modules for map feature extraction.}\vspace{-0.2cm}
\label{tab:cnn_sensitivity}
\setlength\tabcolsep{0.18cm}
\begin{tabular}{>{\hspace{-0.45pc}}ccc<{\hspace{-0.25pc}} cccc c cccc c cccc}
    \toprule
    &&& \multicolumn{4}{c}{\textit{nuScenes}} && \multicolumn{4}{c}{\textit{Lyft}} && \multicolumn{4}{c}{\textit{Waymo}}\\
    \cmidrule(lr){4-7}
    \cmidrule(lr){9-12}
    \cmidrule(lr){14-17}
    & {\footnotesize Infer.} & {\footnotesize \# of} & \multicolumn{2}{c}{minADE$_{k}$ (m)} & \multicolumn{2}{c}{minFDE$_{k}$ (m)} && \multicolumn{2}{c}{minADE$_{k}$ (m)} & \multicolumn{2}{c}{minFDE$_{k}$ (m)} && \multicolumn{2}{c}{minADE$_{k}$ (m)} & \multicolumn{2}{c}{minFDE$_{k}$ (m)} \\
    {\footnotesize CNN Module} & {\footnotesize Time} & {\footnotesize Params.} & $k=1$ & $k=5$ & $k=1$ & $k=5$ && $k=1$ & $k=5$ & $k=1$ & $k=5$ && $k=1$ & $k=5$ & $k=1$ & $k=5$ \\
    \midrule
    ResNet18 & {0.029s} & 11.7M
    & \textbf{3.543} & \textbf{1.586} & \textbf{8.244} & \textbf{3.277} &
    & 0.247 & 0.165 & 0.548 & 0.324 &
    & 0.600 & 0.306 & 1.532 & 0.709 \\
    ResNet152 & {0.102s} & 60.2M
    & 3.728 & 1.728 & 8.773 & 3.657 &
    & \textbf{0.244} & \textbf{0.164} & \textbf{0.544} & \textbf{0.321} &
    & 0.598 & 0.306 & 1.530 & 0.708 \\
    EfficientNet-B0 & {0.029s} & \phantom{0}5.3M
    & 3.715 & 1.740 & 8.733 & 3.676 &
    & 0.246 & 0.165 & 0.548 & 0.325 &
    & 0.585 & 0.299 & 1.491 & 0.686 \\
    MobileNet-V2 & {0.025s} & \phantom{0}3.5M
    & 3.780 & 1.738 & 8.888 & 3.671 & 
    & 0.251 & 0.165 & 0.560 & 0.321 &
    & \textbf{0.585} & \textbf{0.298} & \textbf{1.492} & \textbf{0.684} \\
    \bottomrule
\end{tabular}
\vspace{-0.6cm}
\end{center}
\end{table*}

\subsection{{Computational Performance}}
\label{sec:cnns}

We evaluated the prediction performance of ContextVAE with 
several popular CNN modules for map encoding. 
In Table~\ref{tab:cnn_sensitivity}, we report 
the corresponding results along with the number of parameters and inference time taken by each CNN module.
The inference time refers to the time 
needed to perform $k=5$ predictions for one batch of {32} input samples.
{We pick a batch size of 32
based on the observed statistics in the three tested datasets, where an ego-vehicle has on average $10\pm 7$ agents within an observation radius of $30$m.}
The reported time is measured on a machine equipped with a V100 GPU and averaged over 10,000 test trials.
As can be seen in the table, 
there are no %
large
differences regarding minADE$_k$/minFDE$_k$ among the tested CNN modules on \textit{Lyft} and \textit{Waymo}. %
We can therefore safely choose a small module, like EfficientNet-B0 %
or MobileNet-V2, for the sake of faster inference. 
In \textit{nuScenes}, the best performance is obtained with ResNet18, but the differences among the CNN modules are larger.  
A possible reason is the small size of the \textit{nuScenes} dataset, which makes the model more sensitive to the choice of CNN modules.
For a complete list of all the tested CNN modules, we refer to the supplementary material.
Overall, using ResNet18 on \emph{nuScenes} and \emph{Lyft} and MobileNet-V2 on \emph{Waymo} can meet the real-time inference requirements,
while guaranteeing high prediction accuracy. 
{While ContextVAE only provides marginalized predictions for each agent, the overall time that
it needs to predict the trajectories for all agents on a scene is less than $30\,$ms. 
This is orders of magnitude faster than existing state-of-the-art approaches 
on the leaderboards of %
motion prediction challenges 
(e.g., PGP~\cite{deo2021pgp} needs 407ms on \emph{nuScenes})
and of the same order with approaches that focus on computational efficiency such as PredictionNet~\cite{kamenev2022predictionnet} at the cost though of prediction quality. 
}

{Besides its real-time inference performance, ContextVAE is also very fast to train due to its lightweight architecture that relies on RNNs to perform encoding/decoding sequentially. 
It takes only about 2 hours to train our \emph{nuScenes} model on a single V100 GPU for 80 epochs. In contrast, Trajectron++ needs around nine and a half hours for the same number of epochs. %
The more complex P2T and Autobots-Ego models require more than 2 and 3 days respectively to finish training.  
We note that other state-of-the-art solutions 
are equally slow to train due to the large number of parameters and high complexity of their models, involving transformer-based architectures~\cite{qcnet,shi2022mtr}, multi-stage training phases~\cite{deo2021pgp,liu2023laformer}, ensemble training~\cite{varadarajan2022multipath++,gilles2022gohome}, etc.  %
While such solutions may slightly outperform ContextVAE, they are typically finetuned to a specific dataset by leveraging a number of post-processing techniques and hard-coded assumptions %
In contrast, %
our approach can achieve high-fidelity predictions in real-time for a range of datasets.  
}

\vspace{-8pt}
\subsection{Sensitivity Analysis}\label{sec:sensitivity}
\begin{figure}[!t]
\vspace{-0.2cm}
\centerline{\includegraphics[width=\linewidth]{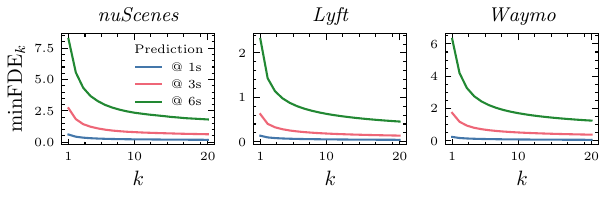}}
    \vspace{-0.4cm}
    \caption{Minimal final displacement error (in meters) as a function of the number $k$ of predicted trajectories for different prediction horizons.}
    \label{fig:error_k}
    \vspace{-0.3cm}
\end{figure}
\paragraph{Prediction Horizon and Choice of $k$}
Intuitively, increasing the prediction horizon leads to more uncertainty, making prediction harder.
This is consistent with the results shown in Fig.~\ref{fig:error_k},  where larger horizons lead to higher minFDE values. %
For a given horizon, an important question is how many prediction samples $k$ are enough to lead to a sufficiently low error? 
For a 1s prediction horizon (blue line), the error converges around $k=10$.
For a 3s horizon (red line), $k=20$ roughly leads to convergence.
However, for a 6s horizon (green line), there is still a decreasing trend at $k=20$. 
We refer to the supplementary material for more 
numeric results of model performance with different prediction horizons.

\paragraph{Model Ablation}
In Table~\ref{tab:ablation},
we perform ablation studies on \emph{Waymo}, related to our proposed unified
observation encoding scheme and its underlying dual attention mechanism (M-ATTN in Eq.~\ref{eq:m_attn} and S-ATTN in Eq.~\ref{eq:s_attn}). 
When neither map nor social context is exploited, the neighbors' states are simply added together for observation encoding. 
When only S-ATTN is used, 
the observation encoder $\mathbf{q}_i^1$ is initialized without map features $\mathbf{M}_i^1$ but social attention is introduced~(Eq.~\ref{eq:s_attn}). 
When the map is introduced independently (``Indie''), 
$\mathbf{M}_i^1$ is %
concatenated with the output $\mathbf{q}_i^T$ of the encoder instead of being used to initialize $\mathbf{q}_i^1$.
In our proposed observation encoding scheme, the map features are integrated into the encoder (``Integrated'') where $\mathbf{q}_i^1$ is initialized with $\mathbf{M}_i^1$ by concatenation~(without M-ATTN) or by the attention mechanism in Eq.~\ref{eq:q1}~(with M-ATTN).
As can be seen from Table~\ref{tab:ablation}, 
S-ATTN brings about a 4\% improvement over the basic timewise VAE backbone.
Handling the semantic map independently from the social context brings an extra 3\% improvement on the deterministic predictions
and around 15\% on the multimodal predictions.
Integrating the map into the 
encoder without M-ATTN 
results in even better performance, highlighting its potential over Indie maps. 
M-ATTN further boosts the performance with an improvement around 35\% on minADE$_5$/minFDE$_5$, and around 15\% on minADE$_1$/minFDE$_1$ compared to the basic backbone architecture.
{Similar conclusions can be drawn from 
\textit{nuScenes} and \textit{Lyft},
as described in the
supplementary material}.

\begin{table}[t]
\begin{center}
\vspace{-0.2cm}
\caption{Model Ablation on \textit{Waymo}.}\vspace{-0.2cm}
\label{tab:ablation}
\begin{tabular}{ccc cccc}
   \toprule
   & & & \multicolumn{2}{c}{minADE$_{k}$ (m)} & \multicolumn{2}{c}{minFDE$_{k}$ (m)} \\
   \scriptsize{S-ATTN} & \scriptsize{MAP} & \scriptsize{M-ATTN} & $k=1$ & $k=5$ & $k=1$ & $k=5$ \\
   \midrule
   - & - & -
   & 0.68 & 0.45 & 1.78 & 1.11 \\
   \ding{51} & - & -
  & 0.65 & 0.43 & 1.71 & 1.06 \\
   \ding{51} & Indie & - 
   & 0.63 & 0.37 & 1.68 & 0.85 \\
   \ding{51} & Integrated & - 
   & 0.62 & 0.33 & 1.65 & 0.70 \\
   \ding{51} & Integrated & \ding{51}
   & \textbf{0.59} & \textbf{0.30} & \textbf{1.49} & \textbf{0.68} \\
   \bottomrule
\end{tabular}

\end{center}
\vspace{-0.3cm}
\end{table}

\begin{figure}[!t]
\vspace{-0.2cm}
\centering
    \includegraphics[width=.9\linewidth]{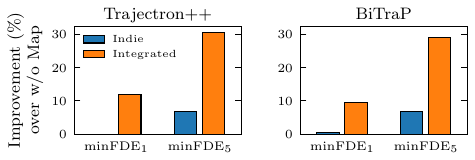} 
    \vspace{-0.3cm}
    \caption{Performance improvement when using map context on \textit{Waymo} dataset. %
    }
    \label{fig:indie_unified}
    \vspace{-0.3cm}
\end{figure}

\begin{figure*}[!t]
    \setlength{\fboxsep}{0pt}
    \centering

    \hfill
    \includegraphics[width=.4\linewidth]{images/scenario_legend.pdf}
    \\\vspace{0.02cm}
    
    \fcolorbox{gray}{white}{\includegraphics[width=.16\linewidth]{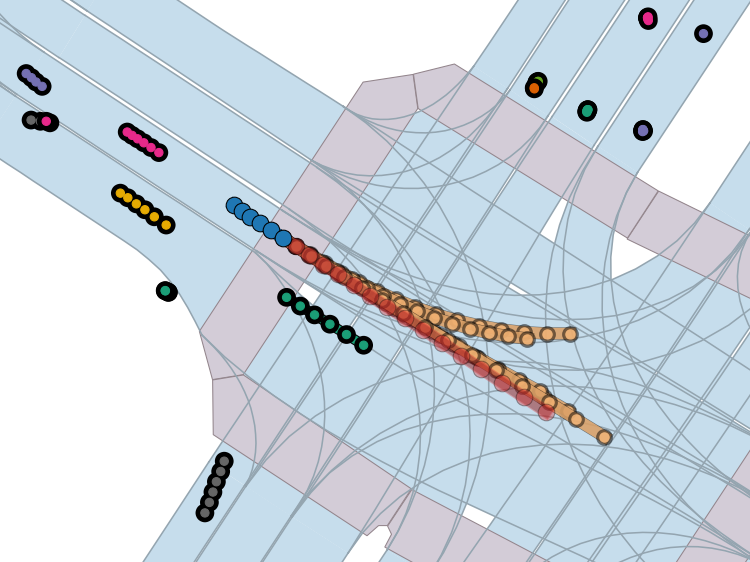}}
    \fcolorbox{gray}{white}{\includegraphics[width=.16\linewidth]{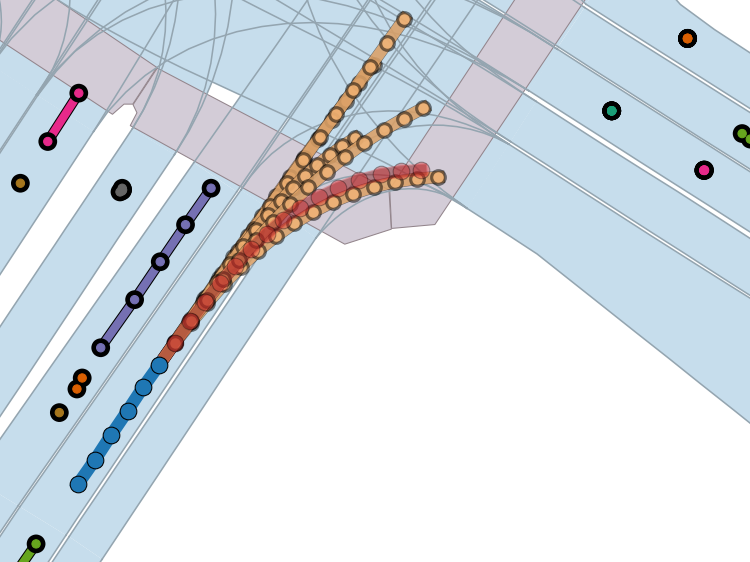}}
    \fcolorbox{gray}{white}{\includegraphics[width=.16\linewidth]{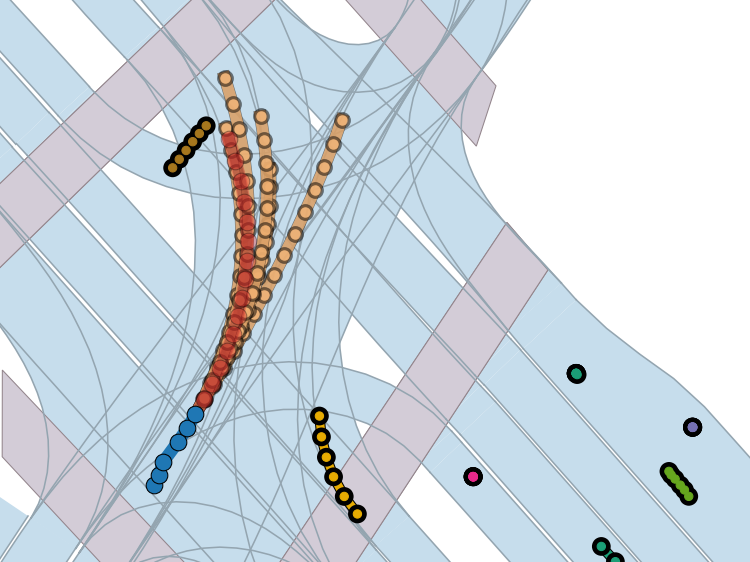}}
    \fcolorbox{gray}{white}{\includegraphics[width=.16\linewidth]{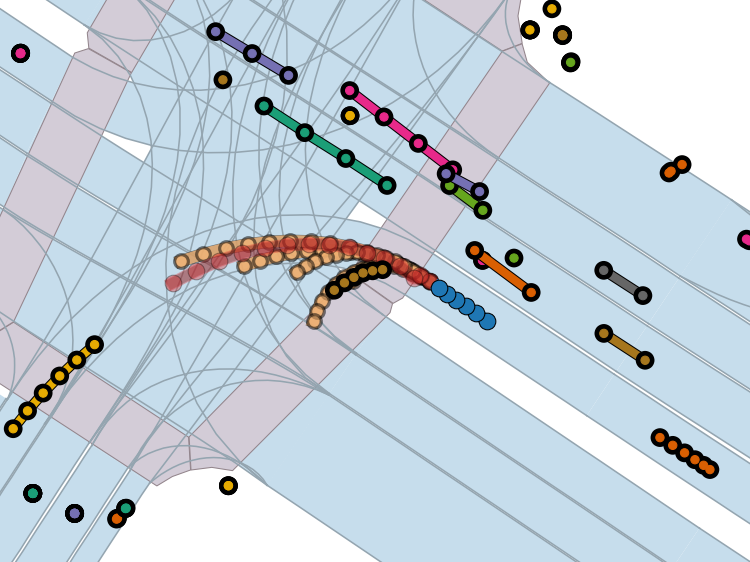}}
    \fcolorbox{gray}{white}{\includegraphics[width=.16\linewidth]{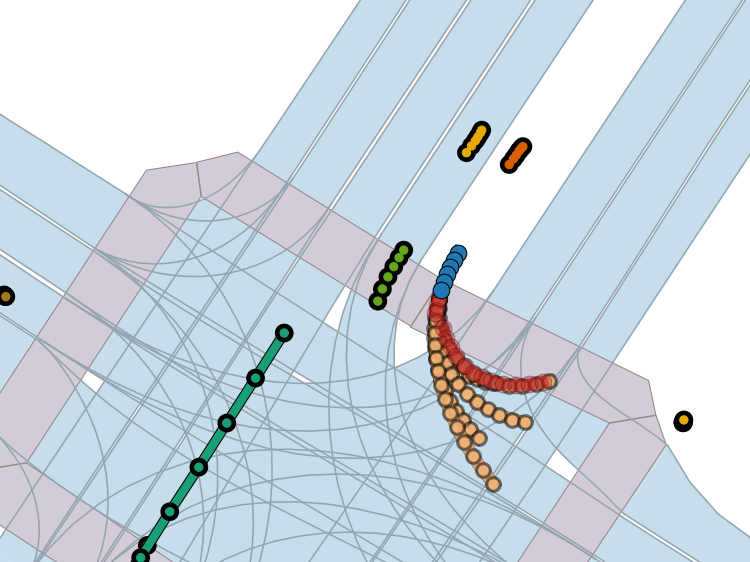}}
    \fcolorbox{gray}{white}{\includegraphics[width=.16\linewidth]{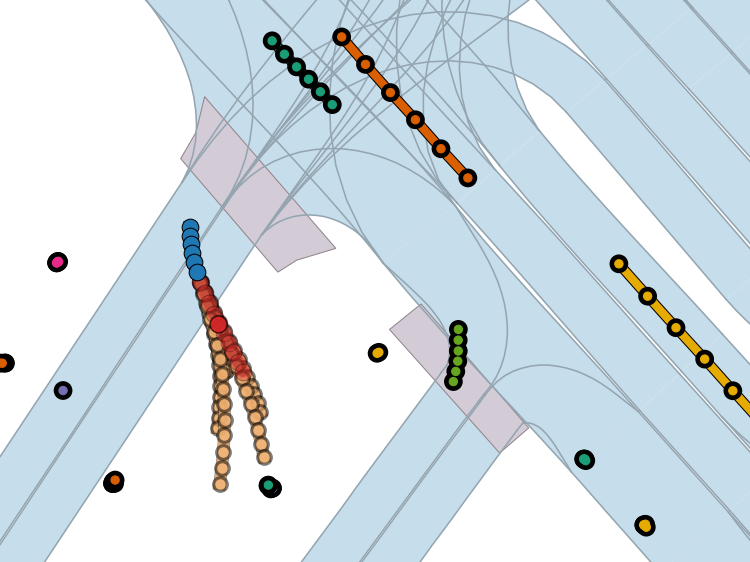}}
    \\\vspace{0.1cm}
    
    \vspace{-0.3cm}
    \caption{Qualitative examples from \textit{Lyft}. Observed trajectories are shown in blue. The ground-truth future trajectories are shown in red. 5 predictions are shown in orange.
    Other colored dots and lines denote stationary and moving neighbors respectively.
    }
    \label{fig:cases} 
    \vspace{-0.3cm}
\end{figure*}
\begin{figure}[!t]
    \setlength{\fboxsep}{0pt}
    \centering
    \begin{flushleft}\scriptsize Prediction with S-ATTN + 
 M-ATTN\end{flushleft}\vspace{-0.1cm}
    
   \fcolorbox{gray}{white}{\includegraphics[width=.32\linewidth]{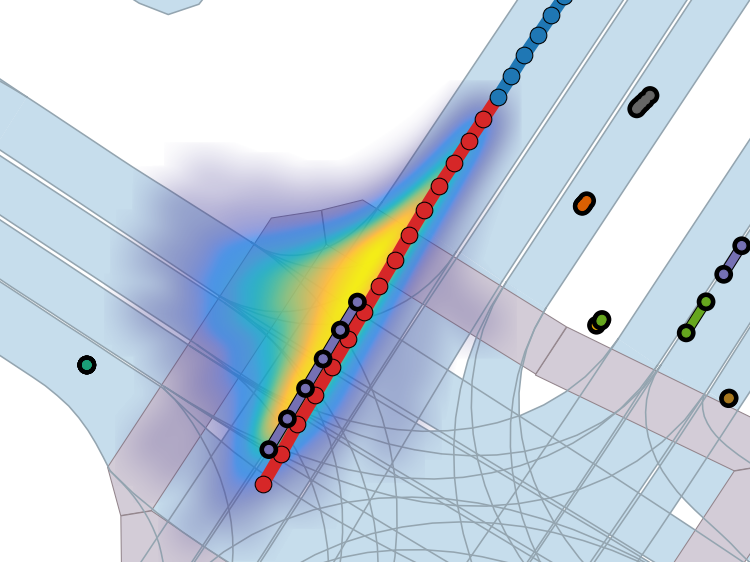}}
  \fcolorbox{gray}{white}{\includegraphics[width=.32\linewidth]{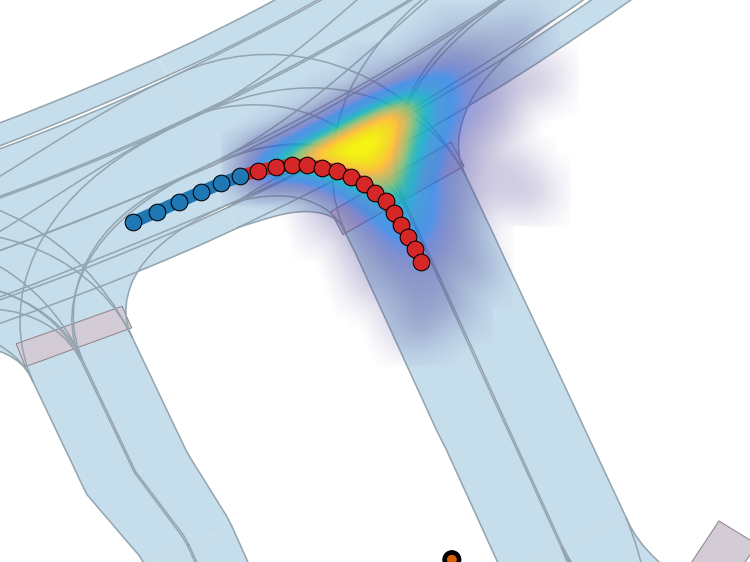}}
   \fcolorbox{gray}{white}{\includegraphics[width=.32\linewidth]{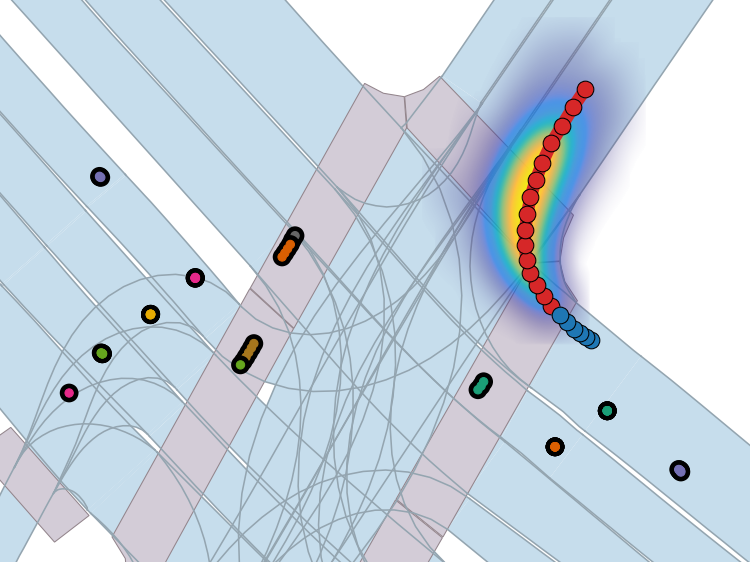}}
    \\\vspace{-0.15cm}
    
    \begin{flushleft}\scriptsize Prediction w/o Map (S-ATTN only)
    \end{flushleft}\vspace{-0.1cm}
    
   \fcolorbox{gray}{white}{\includegraphics[width=.32\linewidth]{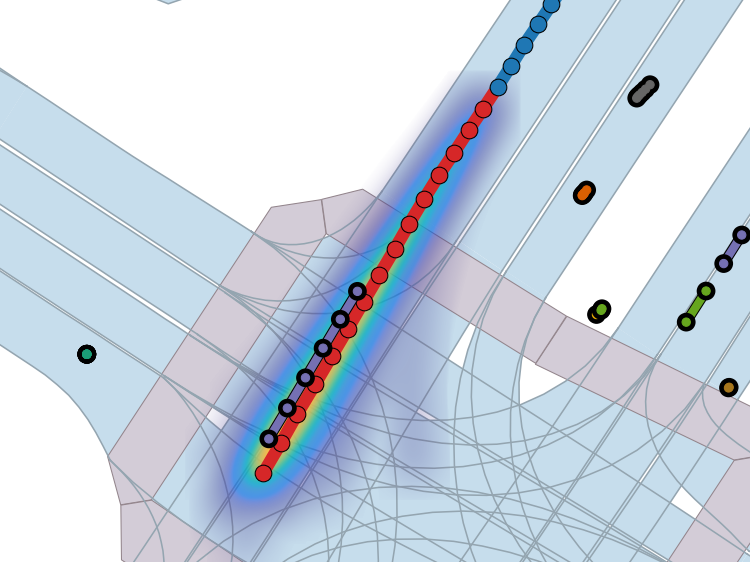}}
   \fcolorbox{gray}{white}{\includegraphics[width=.32\linewidth]{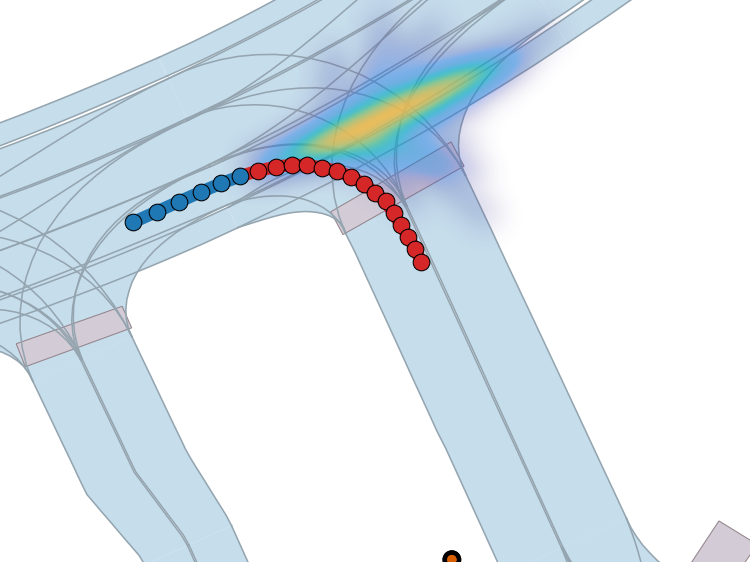}}
  \fcolorbox{gray}{white}{\includegraphics[width=.32\linewidth]{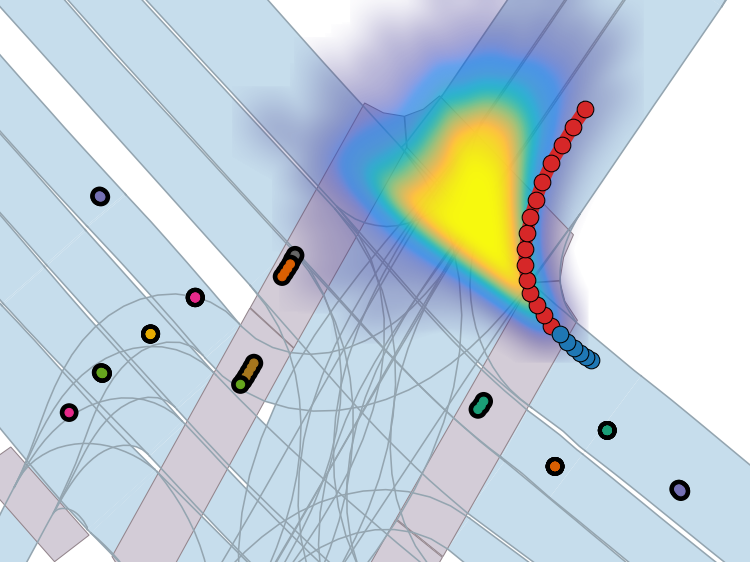}}
    \vspace{-0.5cm}
    \caption{Case studies with (top) and without (bottom) considering rasterized maps in the observation encoder. The trajectory distributions heatmaps are generated using 2,000 predictions {with a Gaussian KDE method.}
    }
    \label{fig:cases_map}  
    \vspace{-0.3cm}
\end{figure}

\subsection{Integrated vs. Independent Maps for Observation Encoding}
\label{sec:indie_vs_uniform}
To further highlight the effectiveness of using Integrated over %
Indie maps for RNN-based observation encoding,  
we consider experiments with additional VAE-based approaches that by default decouple environmental  %
from social context.
Figure~\ref{fig:indie_unified} 
shows the corresponding results for 
Trajectron++ and BiTraP~\cite{yao2021bitrap}. 
As can be seen,
models with Integrated maps significantly improve the  performance of Trajectron++ and Bitrap on both
deterministic and multimodal predictions. In contrast, Indie maps do not provide any noticeable performance gain on deterministic predictions and  only bring a small improvement on multimodal ones.  
We refer to the supplementary material for complete results on all the tested datasets.

\vspace{-9pt}
\subsection{Case Studies}\label{sec:case_study}
Figure~\ref{fig:cases} shows predictions %
obtained with ContextVAE on the \textit{Lyft} dataset. 
As can be seen, 
using maps makes the model aware of the lanes and results in covering 
the ground-truth trajectories closely, while also accounting for other possible driving maneuvers.
In the first three scenes, ContextVAE accurately assigns samples to different lanes at the multi-lane intersections.   
In the fourth, %
it predicts the target vehicle's turning-left behavior along with a potential U-turn behavior. 
The last two cases exemplify non map-compliant behaviors, as the vehicle drives from and to the off-map region, respectively. 
While graph-based methods typically struggle with off-road cases, ContextVAE accurately predicts the vehicles' motion.

In Fig.~\ref{fig:cases_map}, 
we highlight additional examples comparing the prediction results of ContextVAE %
when the map and its related attention mechanism are disabled from the observation encoder. %
Ignoring maps can lead to myopic as well as wrong predictions. 
In the first two cases, the model predicts that the target vehicle will keep driving straight and does not account for the multi-modal decisions at intersections.  
In the third scene, the model predicts multi-modal  trajectories but are invalid as the vehicle can only take a right turn.
In contrast, accounting for map attention leads to map-compliant paths and helps to predict potential turning behaviors %
 at intersections. 

To highlight how our unified observation encoding scheme (Section~\ref{sec:encoding}) helps the model make accurate predictions, we consider three urban traffic examples. In Fig.~\ref{fig:teaser}, we visualize the corresponding activation maps of the map encoder obtained using the GradCAM technique~\cite{gradcam} along with the social attention weights of the perceived neighbors. 
In the first example, the model focuses on the challenging intersection area around the target vehicle and pays high attention to the leading car along its lane. 
In the second example, the most salient map area %
is the crosswalk (light red polygons) 
in front of the target vehicle, where a number of agents are waiting for the traffic lights to turn green. 
Hence, the vehicle's attention focuses on these neighbors while ignoring irrelevant agents such as the three stopped cars on the outer right lane. 
In the third example, the target vehicle is static and the model relies only on the observed neighbors'  states to make predictions with little attention paid to the map. 
These examples demonstrate that our approach can flexibly utilize the environmental map information and the agents' social states to make predictions.

\vspace{-5pt}
\section{Conclusion}
We introduce ContextVAE, a %
{real-time} 
for context-aware vehicle trajectory prediction. 
ContextVAE relies on a timewise VAE architecture and employs 
a map encoding module that performs observation encoding in a unified and socially-aware way. 
We show that our approach provides high-fidelity, map-compliant predictions on a variety of heterogeneous datasets, capturing the multi-modal nature of vehicle motions and their interactions with neighbor agents. 
{We note that the model performance can be further improved by leveraging post-processing techniques like %
clustering approaches~\cite{deo2020p2t,socialvae2022},  ensemble models~\cite{varadarajan2022multipath++,gilles2022gohome},
and other recently proposed schemes~\cite{kamenev2022predictionnet,gilles2021home},  
at the cost of higher running times.}
Interesting avenues for future work are to account for dynamic environmental information like traffic lights in the observation encoding scheme by introducing local maps timewisely, {and 
sampling 
trajectories from the \emph{joint} distribution over a full set of agents
through interactive trajectory prediction~\cite{espinoza2022deep,sun2022m2i}}. %
It is also worth exploring how map representation influences context encoding and model performance,  
including trying alternative ways of map rasterization and using vectorized road graphs. 
{As it stands, our proposed dual-attention mechanism for RNN-based observation encoding enables high-fidelity predictions with low latency, adding value to existing %
approaches for map and agent state fusion. 
Combined with a timewise VAE architecture, the resulting ContextVAE approach can serve as a strong baseline 
for real-time vehicle trajectory prediction and 
inspire future work %
on VAE-based models. 
}

\bibliography{main.bib}
\bibliographystyle{IEEEtran}



\section*{Supplementary material}

TABLE S4
Sensitivity Analysis of Map Encoding Modules on other VAE Models.

| | nuScenes | | | | Lyft | | | | Waymo | | | |
|---|---|---|---|---|---|---|---|---|---|---|---|---|
| | minADE$_k$ | | minFDE$_k$ | | minADE$_k$ | | minFDE$_k$ | | minADE$_k$ | | minFDE$_k$ | |
| | $k = 1$ | $k = 5$ | $k = 1$ | $k = 5$ | $k = 1$ | $k = 5$ | $k = 1$ | $k = 5$ | $k = 1$ | $k = 5$ | $k = 1$ | $k = 5$ |
| Trajectron++ (No map) | 3.97 | 2.50 | 9.24 | 5.89 | 0.37 | 0.31 | 0.88 | 0.62 | 0.88 | 0.70 | 2.37 | 1.60 |
| Trajectron++ (Indie) | 4.08 | 2.41 | 9.67 | 5.63 | 0.38 | 0.26 | 0.89 | 0.57 | 0.88 | 0.56 | 2.37 | 1.41 |
| Trajectron++ (Integrated) | 3.88 | 2.40 | 9.17 | 5.21 | 0.30 | 0.21 | 0.78 | 0.49 | 0.85 | 0.52 | 2.21 | 1.11 |
| BiTraP (No map) | 4.10 | 2.50 | 9.71 | 5.90 | 0.36 | 0.31 | 0.83 | 0.61 | 0.88 | 0.62 | 2.36 | 1.55 |
| BiTraP (Indie) | 4.10 | 2.40 | 9.70 | 5.60 | 0.36 | 0.25 | 0.83 | 0.55 | 0.88 | 0.55 | 2.35 | 1.40 |
| BiTraP (Integrated) | 3.87 | 2.40 | 9.15 | 5.17 | 0.30 | 0.20 | 0.69 | 0.40 | 0.85 | 0.52 | 2.20 | 1.10 |

take the minimal values of ADE and FDE among the $k$ predictions:

$$\text{minADE}_k = \min_{l \in [1,k]} \frac{1}{H} \sum_{t=T+1}^{T+H} \|\mathbf{x}_i^t - \hat{\mathbf{x}}_i^{t,(l)}\|_2. \quad (3)$$

$$\text{minFDE}_k = \min_{l \in [1,k]} \|\mathbf{x}_i^{T+H} - \hat{\mathbf{x}}_i^{T+H,(l)}\|_2. \quad (4)$$

Here, the index $(l)$ refers to the $l$-th trajectory sample (among $k$) realized from the predictive distribution. Note that these metrics do not measure *directly* the divergence between the predictive distribution produced by our system and the true distribution (because we do not have access to this true distribution), but they produce an indirect measure on how close they are.

## II. Semantic Map Rasterization

We use a rasterized, 224×224 local semantic map with a 1:1 ratio between pixels and world coordinates in meters. We refer to Fig. 3 for examples of the rasterized semantic maps used in each test dataset. Because each dataset has its own map data format, the semantic maps used for each model are somewhat different. In *nuScenes* and *Lyft*, the 1st and 2nd channels contain the road divider and lane divider, respectively. The 3rd channel of the *nuScenes* maps consists of the drivable area and the pedestrian crosswalks, obtained through the map API. The drivable area covers lane regions plus some off-road regions where vehicles can move. For *Lyft*, there is no drivable area provided directly by its map API. As such, we draw the drivable areas by filling the provided boundaries of each lane segment. Similarly, *Waymo* provides only road edges as a collection of single lines without the boundary information defined for lane segments. We, therefore, draw the lane center lines across all channels to enhance the representation of drivable areas. Road dividers and edges are drawn on the 1st and 2nd channels respectively, with the crosswalk regions still on the 3rd channel.

## III. Additional Results

### A. Sensitivity Analysis

*a) CNN Modules:* Table S1 shows the model performance when different CNN modules are employed for map feature extraction. Overall, there are no significant differences across all the tested CNN modules on large *Lyft* and *Waymo* datasets. Thus, a small module like MobileNet-V2 or EfficientNet-B0 can be safely used. For *nuScenes*, there is a larger error variance between the different CNN modules with the efficient ResNet18 providing the best performance.

*b) Prediction Horizon:* We analyze the effect that different prediction horizons have on the performance of ContextVAE. We optimize each tested model using the corresponding horizon and report the corresponding minADE/minFDE performance in Table S2.

*c) Ablation Study:* In Table S3, we report the full ablation study results of ContextVAE on all three datasets. The studied components are the same as the ablation study in the main text (cf. Section IV.C). As can be seen, our proposed unified observation scheme that leverages a dual attention mechanism leads to the best performance on all tested datasets.

### B. Integrated vs. Independent Maps for Observation Encoding

Following the analysis in Section IV-E, in Table S4, we report the performance of Trajectron++ [1] and BiTraP [2] when different map integration schemes are employed. Both approaches leverage the same RNN-based observation encoder and employ the same conditional VAE backbone proposed in [1]. When no map is used, the models rely only on the social attention mechanism of Trajectron++. We note that, as opposed to the S-ATTN block of ContextVAE, the social attention of Trajectron++ accounts for the agent's and its neighbors' local states without leveraging any social features. Similar to our ablation studies, when using Indie maps, the observation encoder only accounts for the agent's states and its output $\mathbf{q}_t^i$ is concatenated with the extracted map features $\mathbf{M}_t^i$ before it is passed to the VAE decoder. When Integrated maps are used, we use the map features to initialize the observation encoder, i.e. $\mathbf{q}_t^i = \text{Concat}(\mathbf{M}_t^i, \sum_j f_{\mathbf{n}^1}(\{\mathbf{n}_{j|i}^1\}))$. In both approaches, we use the default map encoding module from Trajectron++ to perform feature extraction. As can be seen in the table, the introduction of Indie maps slightly improves the multimodal performance ($k = 5$) over models that do not rely on semantic maps. However, there is no evident improvement on the most likely, deterministic prediction ($k = 1$). This is consistent with the ablation study results reported in [1]. Using Integrated maps, however, can bring effective improvements to both deterministic and multimodal predictions in all the tested datasets.

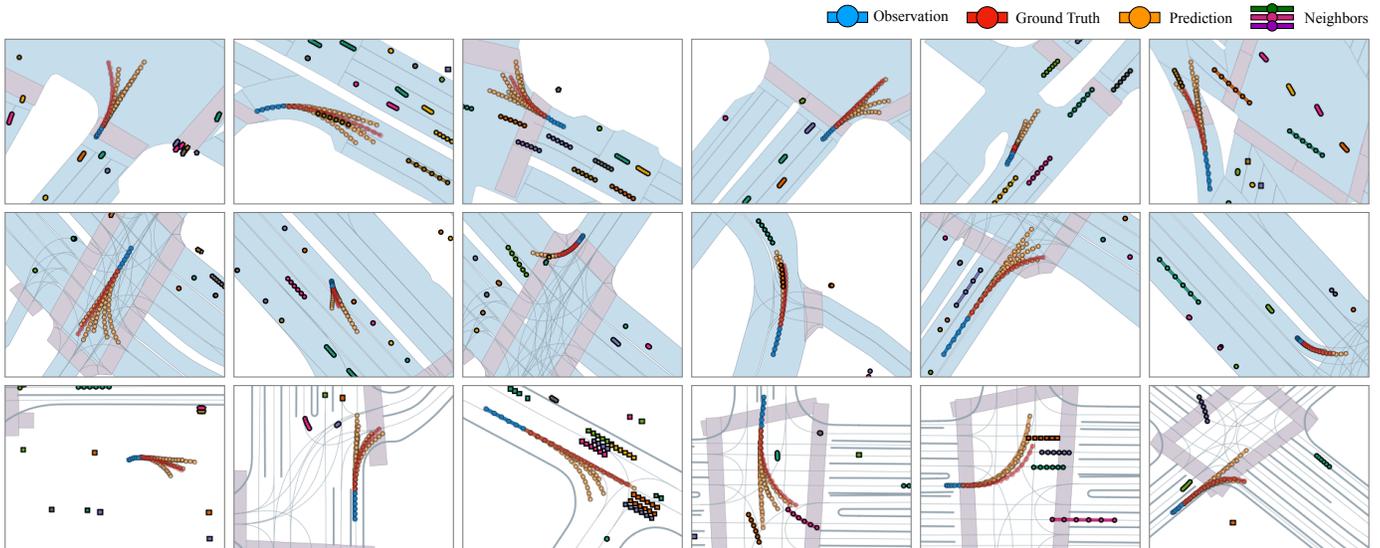

Fig. S1. Examples of ContextVAE predictions on *nuScenes*, *Lyft* and *Waymo* datasets, from top to bottom.

Prediction with S-ATTN + M-ATTN

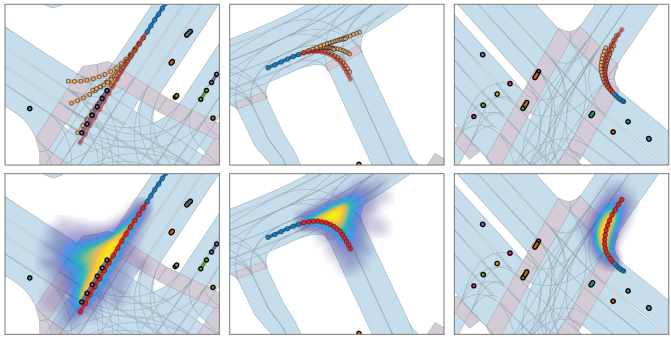

Prediction w/o Map (S-ATTN only)

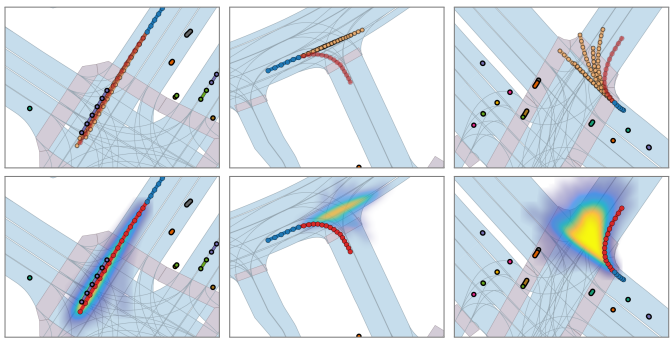

Fig. S2. Case studies with (top) and without (bottom) considering rasterized maps in the observation encoder. The trajectory distributions heatmaps are generated using 2,000 predictions with a Gaussian KDE method.

## IV. CASE STUDIES

In Fig. S1, we show additional examples of our approach on the three tested datasets. Moreover, In Fig. S2, we give a more detailed version of Fig. 7 of the main text, with individual trajectory samples in addition to the heatmaps. We refer to the supplementary video for related results.

Figure S3 depicts an indicative example where ContextVAE performs prediction with and without using the S-ATTN mech-

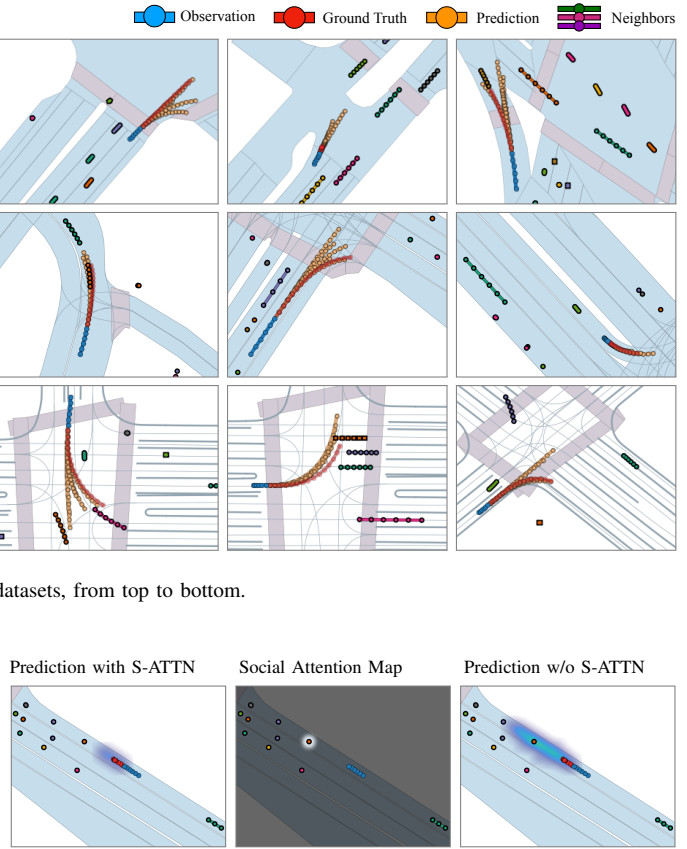

Fig. S3. Case study from *Lyft* with/without accounting for S-ATTN in observation encoding. The left and middle figures show the heatmap of the predicted trajectory distribution and the social attention of the target vehicle, respectively, given neighbors observed within a radius of 30m. The right figure shows the predictions without using the S-ATTN mechanism, i.e., ignoring the pink box from the observation encoder in Fig. 2 of the main text.

anism. As discussed in the ablation study, the social attention mechanism of the observation encoder is equally important, allowing the model to infer the interactions between agents in the scene. In the latter case of Fig. S3, we rely only on the agent's local state $s_i^t$ to sequentially update the RNN hidden state $q_i^t$ of the observation encoder after it has been initialized using our M-ATTN mechanism. Our S-ATTN mechanism pays attention to the stopped car in front of the target vehicle and helps the model to make predictions that are compliant with human driving behavior. Note how the predictive trajectory distribution accurately captures the motion pattern of the target vehicle that slows down to avoid colliding with its neighbor as opposed to when no social attention is used.

In Fig. S4, we qualitatively compare the prediction results provided by our approach to those obtained by the two tested heuristic models of *Constant Velocity* and *Constant Lane*. Compared to *Constant Velocity* where the target vehicle is considered to move with its last observed velocity, *Constant Lane* assumes that the target vehicle will move at its last observed speed but strictly adhere to the lane where it was observed in the last frame; if there is no lane nearby, the target vehicle will be considered to move as *Constant Velocity*. *Constant Lane* ensures that the predicted trajectories are always distributed along the lanes if possible. However, it ignores the possibility

| Constant Velocity | Constant Lane | ContextVAE (Ours) |
|---|---|---|

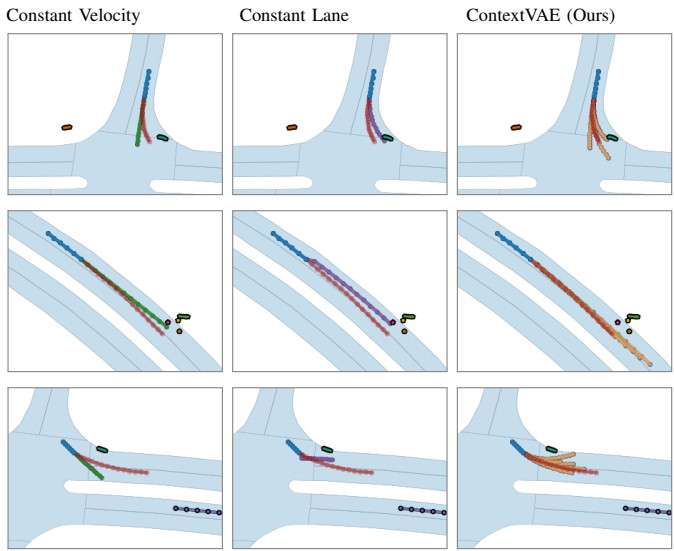

Fig. S4. Qualitative comparison between our approach (right) and the heuristic models of Constant Velocity (left) and Constant Lane (middle).

that the target vehicle may intend to change its lane when its velocity direction differs from the lane direction and also ignores any speed adaptations (the 3rd row of Fig. S4). In addition, both of the heuristic models do not take into account the potential impact with nearby traffic agents as shown in the 2nd row of Fig. S4. In contrast, our approach successfully detects the intention of the target vehicle to change its direction to avoid collision with the neighbors in front of it.



\end{document}